\documentclass[review]{elsarticle}

\usepackage{filecontents}

\usepackage{lineno}
\usepackage[unicode]{hyperref}
\modulolinenumbers[5]

\usepackage{graphicx}
\usepackage{soul}
\usepackage{amssymb,amsmath,amsthm}
\usepackage{graphicx}
\usepackage{color}
\usepackage[ruled,linesnumbered,boxed]{algorithm2e}
\usepackage{footmisc} 
\usepackage{mathtools}
\usepackage{soul}
\usepackage{subcaption}
\usepackage{float}
\SetKwComment{Comment}{$\triangleright$\ }{}

\SetKwProg{Dopar}{do parallel}{}{end}
\usepackage{booktabs,siunitx}

\usepackage[table]{xcolor}
\SetKwInput{kwInput}{Input}
\SetKwInput{kwOutput}{Output}
\usepackage{mathtools} 

\newtheorem{mydef}{Definition}
\begin{document}

\begin{frontmatter}

\title{tax2vec: Constructing Interpretable Features from Taxonomies for Short Text Classification}

\address[ijs]{Jo\v{z}ef Stefan Institute, Jamova 39, 1000 Ljubljana, Slovenia}
\address[mps]{Jo\v{z}ef Stefan International Postgraduate School, Jamova 39, 1000 Ljubljana, Slovenia}
\address[ung]{University of Nova Gorica,  Glavni trg 8, 5271 Vipava, Slovenia}

\author[mps,ijs]{Bla\v{z} \v{S}krlj}
\cortext[1]{Corresponding author}
\author[mps,ijs]{Matej Martinc}

\author[ijs]{Jan Kralj}

\author[ijs,ung]{Nada Lavra\v{c}}

\author[ijs]{*Senja Pollak}




\begin{abstract}
The use of background knowledge is largely unexploited in text classification tasks. This paper explores 
word taxonomies as means for constructing new semantic features, which may improve the performance and robustness of the learned classifiers. We propose tax2vec, a parallel algorithm for constructing taxonomy-based features, and demonstrate its use on six short text classification problems: prediction of gender, personality type, age, news topics, drug side effects and drug effectiveness. The constructed semantic features, in combination with fast linear classifiers, tested against strong baselines such as hierarchical attention neural networks, achieves comparable classification results on short text documents. The algorithm's performance is also tested in a few-shot learning setting, indicating that the inclusion of semantic features can improve the performance in data-scarce situations. The tax2vec capability to extract corpus-specific semantic keywords is also demonstrated. 
Finally, we investigate the semantic space of potential features, where we observe a similarity with the well known Zipf's law.

\end{abstract}

\end{frontmatter}


\section{Introduction}

In text mining, document classification refers to the task of classifying a given text document into one or more categories based on its content \cite{Sebastiani:2002:MLA:505282.505283}. Given an input set of labeled text documents, a text classifier is expected to learn to associate the patterns appearing in the documents to the document labels. 
Deep learning approaches \cite{devlin2018bert} have recently become a standard in natural language-related learning tasks, demonstrating good performance on a variety of different classification tasks, including sentiment analysis of tweets \cite{tang2015document} and news categorization \cite{kusner2015word}. Despite achieving state-of-the-art performance on many tasks, deep learning is not yet optimized for situations, where the number of documents in the training set is low or when the documents contain very little text \cite{rangel2017overview}.

Semantic data mining denotes a data mining approach where domain ontologies are used as background knowledge in the data mining process \cite{lawrynowicz2017semantic}.
Semantic data mining approaches have been successfully applied to association rule learning \cite{angelino2017learning}, semantic subgroup discovery \cite{vavpetic_SDM,wordification}, data visualization \cite{Adhikari2014} and text classification \cite{scott1998text}.
Provision of semantic information allows the learner to use features on a higher semantic level, possibly enabling better data generalizations. The semantic information is commonly represented as relational data in the form of taxonomies or 
3
 ontologies. Development of approaches that leverage such information remains a lively research topic in several fields, including biology \cite{8538944,Chang:2015:HNE:2783258.2783296}, sociology \cite{freeman2017research} and natural language processing \cite{wang2017combining}.

This paper contributes to semantic data mining by using
word taxonomies as means for semantic enrichment by constructing new features, with the goal to improve the performance and robustness of the learned classifiers. In particular, it addresses
classification of short or incomplete documents, which is useful in a large variety of text classification tasks. 
Short text is characterized by shortness in the text length, and sparsity in the terms presented, which results in the difficulty in managing and analyzing them based on the bag-of-words representation only. Short texts can be found everywhere, such as search snippets, product reviews and similar \cite{chen2011short}. 
For example, in author profiling, the task is to recognize the author's characteristics such as age or gender \cite{rangel2014overview}, based on a collection of author's text samples. Here, the effect of data size is known to be an important factor, influencing classification performance \cite{rangel2016overview}. A frequent text type for this task are tweets, where a collection of tweets from the same author is considered a single document, to which a label must be assigned. The fewer instances (tweets) per user we need, the more powerful and useful the approach. Learning from only a handful of tweets can lead to preliminary detection of bots in social networks, and is hence of practical importance \cite{chu2012detecting,chu2010tweeting}.
In a similar way, this holds true for nearly any kind of text classification task. For example, for classifying news into a specific topic, using only snippets or titles may be preferred due to non-availability of entire news texts or for increasing the processing speed.  Moreover, in biomedical applications, Gr\"{a}sser et al. \cite{GraBer} tried to predict drug's side effects and effectiveness from patients' short commentaries, while Boyce et al. \cite{boyce2012using} investigated the use of short user comments to assess drug-drug interactions.

It has been demonstrated that deep neural networks in general need a large amount of information in order to learn complex classifiers, i.e. they require a large training set of documents. For example, the recently introduced BERT neural network architecture \cite{devlin2018bert} consisting of many hidden layers was trained on the whole Wikipedia. It was also shown that the state-of-the-art models do not perform well when incomplete (or scarce) information is used as input \cite{cho2015much}. On the other hand, promising results regarding zero-shot \cite{socher2013zero} and few-shot \cite{snell2017prototypical} learning were recently achieved.

This paper proposes a novel approach named \emph{tax2vec}, where semantic information available in taxonomies is used to construct semantic features that can improve classification performance on short texts. In the proposed approach, features are constructed automatically and remain \emph{interpretable}. We believe that tax2vec could help explore and understand how external semantic information can be incorporated into existing (black-box) machine learning models, as well as help to explain what is being learned. 

This work is structured as follows. Following the theoretical preliminaries and the related work necessary to understand how semantic background knowledge can be used in learning, presented in Section~\ref{sec:background}, 
we continue with the description of the proposed tax2vec methodology in Section~\ref{sec:t2v}. In Section~\ref{sec:exp}, we describe the experimental setting used to test the methodology. In Section~\ref{sec:results}, we present the results of experiments, including the evaluation of the qualitative properties of features constructed using tax2vec, and extensive classification benchmark tests.
Section~\ref{sec:qual+inter} discusses the properties of the resulting semantic space and the explainability of the proposed tax2vec algorithm. Implementation and availability of tax2vec is addressed in Section~\ref{sec:availability}. The paper concludes with a summary and prospects for further work in Section~\ref{sec:conclusions}. For completeness, Appendix A includes a detailed description of the Personalized PageRank algorithm, while Appendix B presents an example segmentation of news articles into paragraphs, forming short documents of interest for this study. Finally, Appendix C contains an additional ablation study regarding the impact of feature numbers on the classifier performance.

\section{Background and related work}
\label{sec:background}

In this section we present the theoretical preliminaries and some related work, which served as the basis for the proposed tax2vec approach. We begin by explaining different levels of semantic context and the rationale behind the proposed approach.

\subsection{Semantic context}
Document classification is highly dependent on \emph{document representation}.
In simple bag-of-words representations, the frequency (or a similar weight such as term frequency-inverse document frequency---tf-idf)
of each word or $n$-gram is considered as a separate feature. More advanced representations group words with similar meaning together. Such approaches include Latent Semantic Analysis 
\cite{landauer2006latent}, Latent Dirichlet Allocation 
\cite{blei2003latent}, and more recently 
word embeddings
\cite{mikolov2013efficient}. It has been previously demonstrated that context-aware algorithms significantly outperform the naive learning approaches \cite{cagliero2013improving}. We refer to such semantic context as the \emph{first-level context}.

\emph{Second-level context} can be introduced by incorporating
\emph{background knowledge} (e.g., ontologies)
into a learning task, which can lead to improved
interpretability and performance of classifiers, learned e.g., by rule learning \cite{vavpetic_SDM} or random forests \cite{xu2018ontological}.
In text mining, Elhadad \textit{et al.} \cite{elhadad2018novel} present an ontology-based web document classifier, while Kaur \textit{et al.}  \cite{kaur2018domain} propose a clustering-based algorithm for document classification that also benefits from knowledge stored in the underlying ontologies.
Cagliero and Garza \cite{cagliero2013improving}
present a custom classification algorithm that can leverage taxonomies and 
demonstrate on a case study of geospatial data that such information can be used to improve the learner's classification performance.
Use of hypernym-based features for classification tasks has been considered previously. For example, hypernym-based features were used in rule learning by the Ripper rule learning algorithm  \cite{scott1998text}. Moreover, it was also demonstrated that the use of hypernym-based features constructed from WordNet significantly impacts the classifier performance \cite{mansuy2006evaluating}.

\subsection{Feature construction and selection}
\label{sec:feat}
When unstructured data is used as input, it is common to explore the options of feature construction. Even though recently introduced deep neural network based approaches operate on simple word indices (or byte-pair encoded tokens) and thus eliminate the need for manual construction of features, such alternatives are not necessarily the optimal approach when vectorizing the background knowledge in the form of taxonomies or ontologies. Features obtained by training a neural network are inherently non-symbolic and as such do not present any added value to the developer's understanding of the (possible) causal mechanisms underlying the learned classifier \cite{bunge2017causality,pearl2009causality}. In contrast,  understanding the semantic background of a classifier's decision can shed light on previously not observed second-level context vital to the success of learning, rendering otherwise incomprehensible models easier to understand.

\begin{mydef}[Feature construction]
Given an unstructured input consisting of $n$ documents, 
a feature construction algorithm outputs a matrix $F \in \mathbb{R}^{n \times \alpha}$, where $\alpha$ denotes the predefined number of features to be constructed.
\end{mydef}
\noindent In practical applications, features are constructed from various data sources, including texts \cite{stanczyk2015feature}, graphs \cite{kakisim2018unsupervised,sge2019}, audio recordings and similar data \cite{tomavsev2015hubness}. With the increasing computational power at one's disposal, automated feature construction methods are becoming prevalent. Here, the idea is that given some criterion, the feature constructor outputs a set of features selected according to the criterion. For example, the tf-idf feature construction algorithm, applied to a given document corpus, can automatically construct hundreds of thousands of n-gram features in a matter of minutes on an average of-the-shelf laptop.

Many approaches can thus output too many features to be processed in a reasonable time, and can introduce additional noise, which renders the task of learning even harder. To solve this problem, one of the known solutions is \emph{feature selection}.

\begin{mydef}[Feature selection]
\label{def:selection}
Let $F \in \mathbb{R}^{n \times \alpha}$ represent the feature matrix (as defined above), obtained during automated feature construction. A feature selection algorithm transforms matrix $F$ to a matrix $F'\in \mathbb{R}^{n \times d}$, where $d$ represents the number of desired features after feature selection.
\end{mydef}

Feature selection thus filters out the (unnecessary) features, with the aim of yielding a compact, information-rich representation of the unstructured input. There exist many approaches to feature selection. They can be based on the individual feature's information content, correlation, significance etc. \cite{chandrashekar2014survey}. Feature selection is, for example, relevant in biological data sets, where only a handful of the key gene markers are of interest, and can be identified by assessing the impact of individual features on the target space \cite{hira2015review}.

\subsection{Learning from graphs and relational information}

In this section we briefly discuss the works that influenced the development of the proposed approach. One of the most elegant ways to learn from graphs is by transforming them into propositional tables, which are a suitable input for many down-stream learning algorithms.
Recent attempts to vectorization of graphs include the node2vec \cite{Grover:2016:NSF:2939672.2939754} algorithm for constructing features from homogeneous networks; its extension  metapath2vec \cite{Dong:2017:MSR:3097983.3098036} for heterogeneous networks; its symbolic version SGE \cite{sge2019}; the mol2vec \cite{doi:10.1021/acs.jcim.7b00616} vectorization algorithm for molecular data; the struc2vec \cite{Ribeiro:2017:SLN:3097983.3098061} graph vectorization algorithm based on homophily relations between nodes, and more. All these approaches (apart from SGE) are sub-symbolic, as the obtained vectorized information (embeddings) are not interpretable. Similarly, recently introduced graph-convolutional neural networks also yield local node embeddings, which take node feature vectors into account \cite{kipf2017semi,NIPS2017_6703}.

In parallel to graph-based vectorization, approaches which tackle the problem of learning from relational databases have also been developed. 
Symbolic (interpretable) approaches for this vectorization task, known under the term propositionalization, include  RSD \cite{vzelezny2006propositionalization}, a rule-based algorithm which constructs relational features; and wordification \cite{perovvsek2013wordification},  an approach for unfolding relational databases into bag-of-words representations.  The approach, described in the following sections, relies on some of the key ideas initially introduced in the mentioned works on propositionalization, as taxonomies are inherently relational data structures.

\section{The tax2vec approach}
\label{sec:t2v}
In this section we outline the proposed tax2vec approach. We begin with a general description of classification from short texts, followed by the key features of tax2vec, which offer solutions to some of the currently not well explored issues in text mining.

\subsection{The rationale behind tax2vec}

In general text classification tasks, deep learning approaches have outperformed other classifiers \cite{devlin2018bert}.  However, in classification tasks involving short documents (tweets, opinions, etc.), particularly where the number of instances is low, deep learners are still outperformed by simpler classifiers, such as SVMs \cite{rangel2019overview}. This observation was a motivation for the development of the tax2vec algorithm, proposed in this paper. Compared to non-symbolic node vectorization algorithms discussed in the previous section, tax2vec uses hypernyms as potential features directly and thus makes the process of feature construction and selection possible without the loss of classifier's \emph{interpretability}.

\begin{figure}
    \centering
    \includegraphics[height=0.5\linewidth,width=0.8\linewidth]{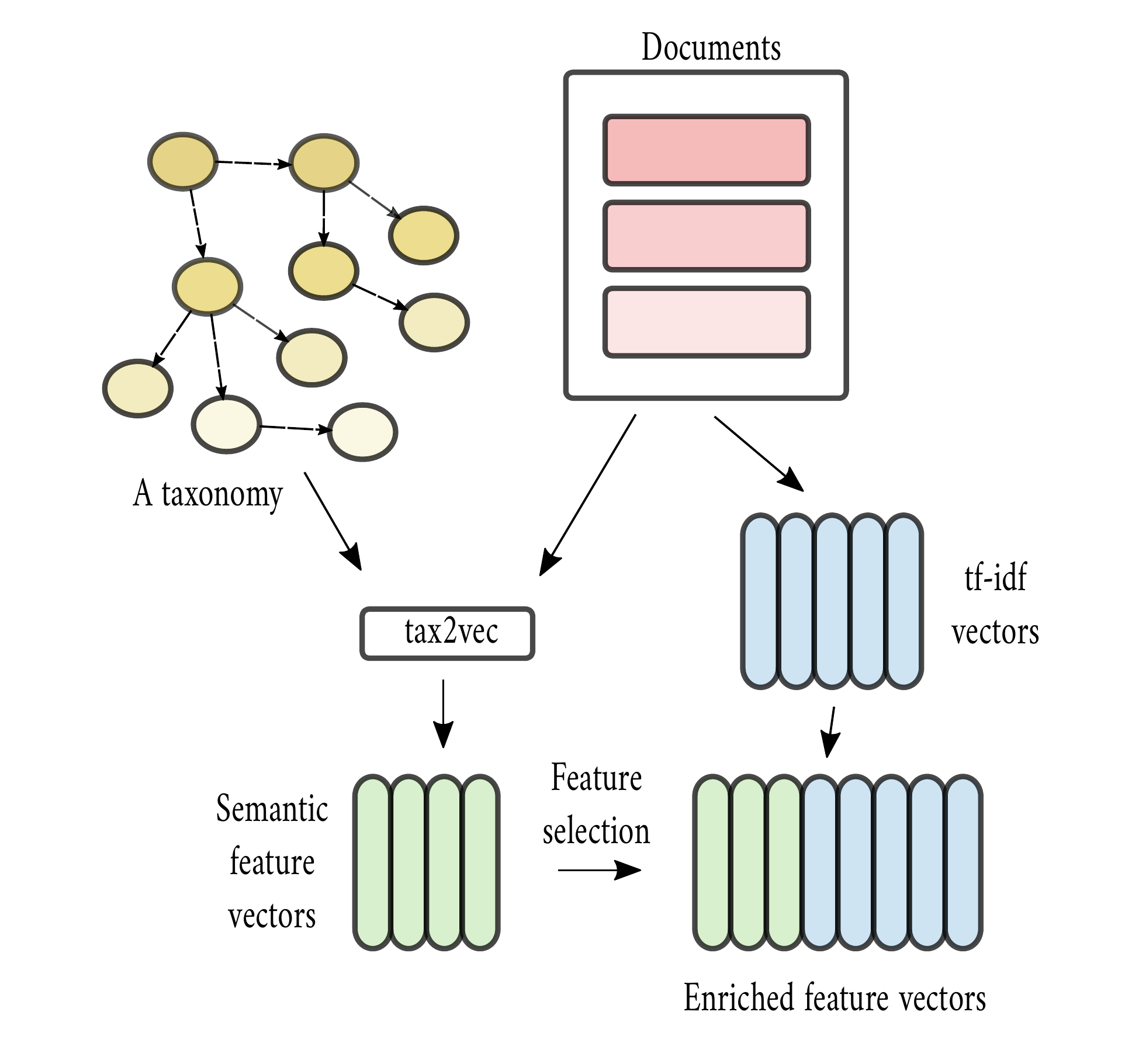}
    \caption{Schematic representation of tax2vec, combined with standard tf-idf representation of documents. Note that darker nodes in the taxonomy represent more general terms.}
    \label{fig:workflow}
\end{figure}

We present the proposed tax2vec algorithm for semantic feature vector construction that can be used to enrich the feature vectors constructed by the established text processing methods such as tf-idf. The tax2vec algorithm takes as input a labeled or unlabeled corpus of $n$ documents and a word taxonomy. It outputs a matrix of \emph{semantic feature vectors} in which each row represents a semantics-based vector representation of one input document. Example use of tax2vec in a common language processing pipeline is shown in Figure~\ref{fig:workflow}. Note that the obtained semantic feature vectors serve as additional features in the final, vectorized representation of a given corpus. 


Let us first explore how parts of the WordNet taxonomy \cite{Miller:1995:WLD:219717.219748} related to the training corpus can be used for the construction of novel features, as such background knowledge can be applied in virtually every English text-based learning setting, as well as for many other languages \cite{Gonzalez-Agirre:Laparra:Rigau:2012}. 

\subsection{Deriving semantic features}
\label{sec:derivingsemanticfeatures}

The tax2vec approach implements a two-step semantic feature construction process. First, a document-specific taxonomy is constructed, then a term-weighting scheme is used for feature construction.

\subsubsection{Document-based and corpus-based taxonomy construction}
\label{subs:local_taxonomy_construction}

In the first step of the tax2vec algorithm, a corpus-based taxonomy is constructed from the input document corpus. In this section we describe how the words from individual documents of a corpus are mapped to terms of the WordNet taxonomy to construct a \emph{document-based taxonomy} by focusing on semantic structures, derived exclusively from the \emph{hypernymy} relation between words. Individual document-based taxonomies are then merged into a joint \emph{corpus-based taxonomy}. 

When constructing a document-based taxonomy, each word is mapped to the hypernym WordNet taxonomy. This results in a tree-like structure, which spans from individual words to higher-order semantic concepts. For example, given the word monkey, one of its mappings in the WordNet hypernym taxonomy is the term \emph{mammal}, which can be further mapped to e.g., \emph{animal} etc., eventually reaching the most general term, i.e. \emph{entity}.

In order to construct the mapping, the first problem to be solved is  \emph{word-sense disambiguation}. For example, the word \textit{bank} has two different meanings, when considered in the following two sentences:

\begin{table}[H]
\centering
\begin{tabular}{c|c}
River bank was enforced. & National bank was robbed.
\end{tabular}
\end{table}
There are many approaches to word-sense disambiguation (WSD). We refer the reader to \cite{Navigli:2009:WSD:1459352.1459355} for a detailed overview of the WSD methodology. 

In tax2vec, we use Lesk \cite{basile2014enhanced}, the gold standard WSD algorithm, to map each disambiguated word to the corresponding term in the WordNet taxonomy. The identified term is then associated with a path in the WordNet taxonomy leading from the given term to the root of the taxonomy. Example hypernym path (with WordNet-style notation), extracted for word ``astatine'', is shown in Figure~\ref{fig-astatine}. 
\begin{figure}[ht]
\indent
$Synset('entity.n.01') \\ \rightarrow Synset('abstraction.n.06') \\ \rightarrow Synset('relation.n.01') \\ \rightarrow  Synset('part.n.01') \\ \rightarrow Synset('substance.n.01') \\ \rightarrow  Synset('chemical\_element.n.01') \\ \rightarrow  Synset('astatine.n.01')$
\caption{Example hypernym path extracted for word ``astatine'', where the $\rightarrow$ corresponds to the ``hypernym of'' relation (the majority of hypernym paths end with the ``entity'' term, as it represents one of the most general objects in the taxonomy).}
\label{fig-astatine}
\end{figure}

By finding a hypernym path to the root of the taxonomy for all words in the input document, a \emph{document-based taxonomy} is constructed, which consists of all hypernyms of all words in the document. After constructing the document-based taxonomy for all the documents in the corpus, the taxonomies are joined into a \emph{corpus-based taxonomy}.

Note that processing each document and constructing the document-based taxonomy is entirely independent from other documents, allowing us to process the documents in parallel and join the results only when constructing the joint corpus-based taxonomy.

\subsubsection{Semantic feature construction}

During the construction of a document-based taxonomy, document-level term counts are calculated for each term. For each word $t$ and document $D$, we count the number $f_{t,D}$ of times the word or one of its hypernyms appeared in a given document $D$.

The obtained counts can be used for feature construction directly: each term $t$ from the corpus-based taxonomy is associated with a feature, and a 
document-level term count is used as the feature value. The current implementation of tax2vec weights the feature values using the double normalization tf-idf metric. For term $t$, document $D$ and user-selected normalization factor $K$,   feature value \textrm{tf-idf}(t,D,K) is calculated as follows \cite{manning}:
\begin{equation}
    \textrm{tf-idf}(t,D,K) = \underbrace{ \bigg (K+(1-K)\frac{f_{t,D}}{\max_{\{t' \in D\}} f_{t',D}} \bigg )}_{\textrm{Weighted term frequency}} \cdot \underbrace{\log\left(\frac{N}{n_t}\right)}_{\substack{\textrm{Inverse}\\ \textrm{document frequency}}}
\end{equation}

\noindent where $f_{t,D}$ is the term frequency, normalized by $\max_{\{t' \in D\}}f(t',D)$, which corresponds to the raw count of the most common hypernym of words in the document; value $N$ represents the total number of documents in the corpus, $n_t$ denotes the number of document-based taxonomies the hypernym appears in (i.e. the number of documents that contain a hyponym of $t$). Note that the term frequencies are normalized with respect to the most frequently occurring term to prevent a bias towards longer documents. In the experiments the normalization constant $K$ was set to $0.5$.  

\subsection{Feature selection}

The problem with the above presented approach is that all hypernyms from the corpus-based taxonomy are considered, and therefore, the number of columns in the feature matrix can grow to tens of thousands of terms. Including all these terms in the learning process introduces unnecessary noise, and unnecessarily increases the spatial complexity. This leads to the need of feature selection (see  Definition~\ref{def:selection} in Section~\ref{sec:feat}) to reduce the number of features to a user-defined number (a free parameter specified as part of the input). We next describe the scoring functions of feature selection approaches considered in this work.

As part of tax2vec, we implemented both supervised (Mutual Information - MI and Personalized PageRank - PPR), as well as unsupervised (Betweenness centrality - BC and term count-based selection) feature selection methods, discussed below. Note that the feature selection process is conducted \emph{exclusively} on the semantic space (i.e. on the mapped WordNet terms). 

\begin{description}
\item[\textbf{Feature selection by term counts.}]
Intuitively, the rarest terms are the most document-specific and could provide additional information to the classifier. This is addressed in tax2vec by the simplest heuristic, used in the algorithm: a term-count based heuristic that simply takes overall counts of all hypernyms in the corpus-based taxonomy, sorts them in ascending order according to their frequency of occurrence and takes the top $d$.

\item[\textbf{Feature selection using term betweenness centrality.}]
As the constructed corpus-specific taxonomy is not necessarily the same as the WordNet taxonomy, the graph-theoretic properties of individual terms within the corpus-based taxonomy could provide a reasonable estimate of a term's importance. The proposed tax2vec implements the betweenness centrality (BC) \cite{brandes} measure of individual terms as the scoring measure. The betweenness centrality is defined as:
\begin{equation}
BC(t) = \sum_{u \neq v \neq t}\frac{\sigma_{uv}(t)}{\sigma_{uv}};
\end{equation}
where $\sigma_{uv}$ corresponds to the number of shortest paths (see Figure~\ref{fig:sp}) between nodes $u$ and $v$, and $\sigma_{uv}(t)$ corresponds to the number of paths that pass through term (node) $t$. Intuitively, betweenness measures the $t$'s importance in the corpus-based taxonomy. Here, the terms are sorted in a descending order according to their betweenness centrality, and again, the top $d$ terms are used for learning.

\begin{figure}[ht]
\centering
\includegraphics[width=0.3\linewidth]{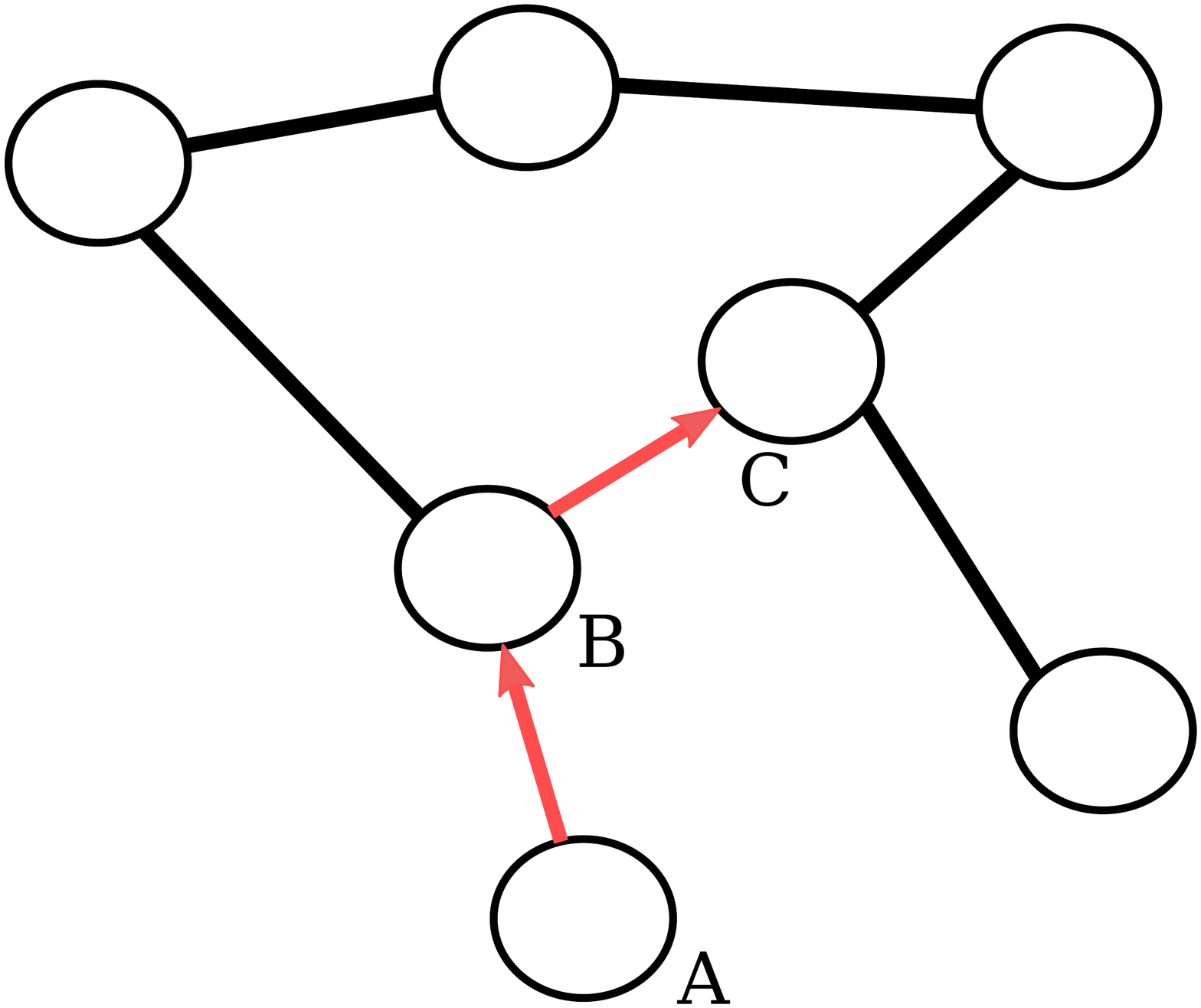}
\caption{An example shortest path. The path colored red represents the smallest number of edges needed to reach node C from node A.} 
\label{fig:sp}
\end{figure}

\item[\textbf{Feature selection using mutual information.}]
The third heuristic, mutual information (MI) \cite{peng2005feature}, aims to exploit the information from the labels, assigned to the documents used for training. The MI between two random discrete variables represented as vectors $F_i$ and $Y$ (i.e. the $i$-th hypernym feature and a target binary class) is defined as:
\begin{equation}
MI(F_i,Y) = \sum_{x,y \in \{0,1\}} p(F_{i} = x,Y=y) \cdot
  \log_2 \bigg(\frac{p(F_{i}=x,Y=y)}{p(F_{i }=x) \cdot p(Y=y)}\bigg )
\end{equation}
\noindent where $p(F_i=x)$ and $p(Y=y)$ correspond to marginal distributions of the joint probability distribution of $F_i$ and $Y$. Note that for this step, tax2vec uses the binary feature representation, where the tf-idf features are rounded to the closest integer value (either 0 or 1). This way, only well represented features are taken into account. Further, tax2vec uses one-hot encodings of target classes, meaning that each target class vector consists exclusively of zeros and ones.
For \emph{each} of the target classes, tax2vec computes the mutual information (MI) between \emph{all} hypernym features (i.e. matrix $X$) and a given class. Hence, for each target class, a vector of mutual information scores is obtained, corresponding to MI between individual hypernym features and a given target class.

Finally, tax2vec sums the MI scores obtained for each target class to obtain the final vector, which is then sorted in descending order. The first $d$ hypernym features are used for learning.
At this point tax2vec yields the selected features as a sparse matrix, maintaining the spatial complexity amounting to the number of float-valued non-zero entries.

\item[\textbf{Personalized PageRank-based hypernym ranking.}]
Advances by Kralj \emph{et al.} \cite{JMLR:v20:17-066, kralj2017heterogeneous} in learning using extensive background knowledge for rule induction explored the use of Personalized PageRank (PPR) algorithm for node subset selection in semantic search space exploration. In tax2vec, we use the same idea to prioritize (score) hypernyms in the corpus-based  taxonomy. In this section, we first briefly describe the Personalized PageRank algorithm and then describe how it is applied in tax2vec.

The PPR algorithm takes as an input a network and a set of starting nodes in the network and returns a vector assigning a score to each node in the input network. The scores of nodes are calculated as the stationary distribution of the positions of a random walker that starts its walk on one of the starting nodes and, in each step, either randomly jumps from a node to one of its neighbors (with probability $p$, set to $0.85$ in our experiments) or jumps back to one of the starting nodes (with probability $1-p$). Detailed description of the PPR used in tax2vec is given in Appendix A. The PPR algorithm is used in tax2vec as follows:

\begin{enumerate}
\item Identify a set of hypernyms in the corpus-based taxonomy, to which the words in the input corpus map to in the first step of tax2vec (described in Section \ref{subs:local_taxonomy_construction}).
\item Run the PPR algorithm on the corpus-based taxonomy, using the hypernyms identified in step $1$ as the starting set.
\item Use the top $d$ best ranked hypernyms as candidate features.
\end{enumerate}

Note that this heuristics offers \emph{global} node ranks with respect to the corpus used.
\end{description}

\subsection{The tax2vec algorithm}

All the aforementioned steps form the basis of tax2vec, outlined in Algorithm~\ref{algo:ttv}. First, tax2vec iterates through the given labeled document corpus in parallel (lines 3--7).  For each document, \emph{MaptoTaxonomy} method identifies a set of disambiguated words and determines their corresponding terms in taxonomy $\mathfrak{T}$ (i.e. WordNet) using method $m$ (i.e. Lesk). Term counts are stored for later use (\emph{storeTermCounts}), and the taxonomy, derived from a given document ($doc$) is added to the corpus taxonomy $\mathfrak{T}_{\textsc{CORPUS}}$.
Once traversed, the terms present in $\mathfrak{T}_{\textsc{CORPUS}}$ represent potential \emph{features}. Term counts, stored for each document are aggregated into vectors of size n, where n is the number of documents in the corpus. 
The result of this step is a real-valued, sparse matrix (vecSpace), where columns represent all possible terms from $\mathfrak{T}_{\textsc{CORPUS}}$. In the following step, feature selection is conducted. Here, graph-based methods (e.g., BC and PPR) identify top $d$ terms based on $\mathfrak{T}_{\textsc{CORPUS}}$'s properties (lines 9--12), and non-graph methods (e.g., MI) is used directly on the sparse matrix to select which $d$ features are the most relevant (lines 13--15). Finally, $selectedFeatures$, a matrix of selected semantic features is returned.

\begin{algorithm}[t!]
\KwData{Training set documents $D$, training document labels $Y_{tr}$, WordNet taxonomy $\mathfrak{T}$, word-to-taxonomy mapping $m$, feature selection heuristic $h$, number of selected features $d$}
$\mathfrak{T}_{\textsc{CORPUS}}$ $\leftarrow$ empty structure\;
termCounts $\leftarrow$ empty structure\;
\For{$doc \in D$ (in parallel)}{
    $\mathfrak{T}_{\textsc{DOCUMENT}}$ $\leftarrow$ MaptoTaxonomy$(doc, \mathfrak{T}, m)$\;
    Add storeTermCounts($\mathfrak{T}_{\textsc{DOCUMENT}}$) to termCounts\;
    Add $\mathfrak{T}_{\textsc{DOCUMENT}}$ to $\mathfrak{T}_{\textsc{CORPUS}}$\;
}
vecSpace $\leftarrow$ tf-idf(constructTfVectors$(D,\mathfrak{T}_{\textsc{CORPUS}},$termCounts$)$)\;
\If{$h$ is graph-based}{
    topTerms $\leftarrow$ selectFeatures(h, $\mathfrak{T}_{\textsc{CORPUS}}$, d, optional $Y_{tr}$)\;
    selectedFeatures $\leftarrow$ select topTerms from vecSpace\;
}\Else{
    selectedFeatures$\leftarrow$ selectFeaturesDirectly(h, vecSpace,d ,$Y_{tr}$)\;
}
\KwRet selectedFeatures\;
\KwResult{$d$ new feature vectors in sparse vector format.}
\caption{tax2vec}
\label{algo:ttv}
\end{algorithm}


Note that in practice, tax2vec must also store the inverse document frequencies in order to generate features for unseen documents. We omit the description of this step for readability purposes.

\subsection{Handling noise}
\label{sec:noise}

Numerous data sets, including contemporary social media data sets, can be noisy and as such hard to handle by a learning system. We next discuss how distinct parts of tax2vec potentially handle noise in the data, including typos, incomplete and missing words and uncommon characters.

During the initial step of the semantic space construction, tax2vec conducts document-level word disambiguation in order to semantically characterize a given token (word). During this step, any tokens that are not present in the taxonomy will be ignored. Further, as word disambiguation requires a certain word window to operate, this hyperparameter can be used to control the size of context considered by tax2vec. In this work, however, we did not explicitly address the problem of invalid tokens in a given token's neighborhood, yet observed that small window sizes (two and three) offered reasonably robust performance.

Even though disambiguation with Lesk offers the initial \emph{semantic pruning} capabilities, the tax2vec algorithm can further address potential noise as follows. As the user can determine the depth in the WordNet taxonomy that will be considered as the starting point for semantic space construction, potentially too specific terms can be avoided if necessary.

Finally, in the third step, tax2vec conducts \emph{feature selection}. This part of the algorithm is responsible for \emph{filtering} redundant and non-informative terms that could be considered as noise. We tested both supervised, as well as unsupervised feature selection methods, exploring whether additional information about class labels helps with term pruning.
Apart from the semantic pruning and selection strategies discussed above, links, mentions and hashtags can be removed to further reduce the noise in social media texts (as mentioned in the description of the SVM implementation by Martinc et al. \cite{Martinc2017PAN2A} in Section \ref{subsec:classifiers}).

We believe all three steps to some extent address how noise is being handled. However, it is expected that additional grammar correction and text normalization could serve as a complementary step to offer improved performance on social media texts.

\section{Experimental setting}
\label{sec:exp}
This section presents the experimental setting used in testing the performance of tax2vec in document classification tasks. We begin by describing the data sets on which the method was tested. Next, we describe the classifiers used to assess the use of features constructed using tax2vec, along with the baseline approaches. We continue by describing the metrics used to assess  classification performance, and the description of the experiments.

\subsection{Data sets}
We tested the effects of features produced with tax2vec on six different class labeled text data sets summarized in Table~\ref{tbl:summary}, intentionally chosen from different domains.

\begin{table}[ht]
\centering
\caption{Data sets used for experimental evaluation of tax2vec's impact on learning. Note that MNS corresponds to the maximum number of text segments (max. number of tweets or comments per user or number of news paragraphs as presented in Appendix B). }
\resizebox{\textwidth}{!}{
\begin{tabular}{l|cccccc}

                       Data set (target) &  Classes &  Words &  Unique words &  Documents &  MNS & Average tokens per segment \\ \hline
    PAN 2017 (Gender) &                  2 &          5169966 &        607474 &                 3600 &                     102 &                   14.23 \\
                 MBTI (Personality) &                 16 &         11832937 &        372811 &                 8676 &                      89 &                   27.98 \\
    PAN 2016 (Age) &                  5 &           943880 &        178450 &                  402 &                     202 &                   13.17 \\
                BBC news &                  5 &           902036 &         58128 &                 2225 &                      76 &                   70.39 \\
     Drugs (Side effects) &                  4 &           385746 &         27257 &                 3107 &                       3 &                   41.47 \\
  Drugs (Overall effect) &                  4 &           385746 &         27257 &                 3107 &                       3 &                   41.47 \\
\hline
\end{tabular}
}
\label{tbl:summary}
\end{table}

The first three data sets are composed of short documents from social media, where we consider classification of tweets.


\begin{description}
\item[\textbf{PAN 2017 (Gender) data set.}] Given a set of tweets per user, the task is to predict the user's gender\footnote{\url{https://pan.webis.de/clef17/pan17-web}} \cite{rangel2017overview}.
    
\item[\textbf{MBTI (Meyers-Briggs personality type) data set.}] Given a set of tweets per user, the task is to predict to which personality class a user belongs\footnote{\url{https://www.kaggle.com/datasnaek/mbti-type/kernels}}, first discussed in \cite{myers1962myers}.
    
\item[\textbf{PAN 2016 (Age) data set.}] Given a set of tweets per user, the classifier should predict the users's age range\footnote{\url{https://pan.webis.de/clef18/pan18-web}} \cite{rangel2016overview}.
\end{description}

Next, we consider a news articles data set by which we test the potential of the method also on longer documents, while for few shot learning experiments (Section~\ref{sec:fsl}), we transform the setting to short text documents by using only few paragraphs per article and test whether competitive performance to full-text-based classification can be obtained.
\begin{description}
    \item[\textbf{BBC news data set.}] Given a news article (composed of a number of paragraphs)\footnote{Split to paragraphs according to the double new line is presented in Appendix B.}, the goal is to assign to it a topic from a list of topic categories\footnote{\url{https://github.com/suraj-deshmukh/BBC-Dataset-News-Classification/blob/master/dataset/dataset.csv}} \cite{greene06icml}.
\end{description}

We also consider two biomedical data sets related to drug consumption. Here, the same training instances in the form of short user commentaries were used to predict two different targets.
\begin{description}
\item[\textbf{Drug side effects.}] This data set links user opinions to side effects of a drug they are taking as treatment. The goal is to predict the side effects prior to experimental measurement \cite{GraBer}.\footnote{\url{http://archive.ics.uci.edu/ml/datasets}}

\item[\textbf{Drug effectiveness.}] Similarly to side effects (previous data set), the goal of this task is to predict drug effectiveness  \cite{GraBer}.
\end{description}


\subsection{The classifiers used}
\label{subsec:classifiers}
As tax2vec serves as a preprocessing method for data enrichment with semantic features, arbitrary classifiers can use the resulting semantic features for learning. Note that in the experiments, the final feature space is composed of both semantic and non-semantic (original) features, i.e., the final feature set used for learning is formed \emph{after} the semantic features have been constructed and selected, by concatenating the original features and the semantic features.
We use the following learners:
\begin{description}
\item[\textbf{PAN 2017 approach.}]
An SVM-based approach that relies heavily on the method proposed by Martinc et al. \cite{Martinc2017PAN2A} for the author profiling task in the PAN 2017 shared task \cite{rangel2017overview}. This method is based on sophisticated hand-crafted features calculated on different levels of preprocessed text including optional social media text cleaning (e.g., Twitter hashtag, mentions, url replacement with filler tokens). The following features were used:

\begin{description}
\item[tf-idf weighted word unigrams] calculated on lower-cased text with stopwords removed;
\item[tf-idf weighted word bigrams] calculated on lower-cased text with punctuation removed;
\item[tf-idf weighted word bound character tetragrams] calculated on lower-cased text;
\item[tf-idf weighted punctuation trigrams] (the so-called beg-punct \cite{SapkotaBMS15}, in which the first character is punctuation but other characters are not) calculated on lower-cased text; 
\item[tf-idf weighted suffix character tetragrams] (the last four letters of every word that is at least four characters long \cite{SapkotaBMS15}) calculated on lower-cased text;
\item[emoji counts] of the number of emojis in the document, counted by using the list of emojis created by \cite{novak2015sentiment}\footnote{http://kt.ijs.si/data/Emoji\_sentiment\_ranking/}; this feature is only useful if the input text contains emojis;
\item[document sentiment] using the above-mentioned emoji list that contains the sentiment of a specific emoji, used to calculate the sentiment of the entire document by simply adding the sentiment of all the emojis in the document; this feature is only useful if the input text contains emojis; 
\item[character flood counts] calculated by the number of times that three or more identical character sequences appear in the document;
\end{description}

In contrast to the original approach proposed  \cite{Martinc2017PAN2A}, we do not use POS tag sequences as features and a Logistic regression classifier is replaced by a Linear SVM. Here, we experimented with the regularization parameter C, for which values in range $\{1,20,50,100,200\}$ were tested. This SVM variant is from this point on referred to as ``SVM (Martinc et al.)''. As this feature construction pipeline consists of too many parameters, we were not able to perform extensive grid search due to computational complexity. Thus, we did not experiment with feature construction parameters, and kept the configuration proposed in the original study. 

\item[\textbf{Linear SVM with automatic feature construction.}]

The second learner is a libSVM linear classifier \cite{chang2011libsvm}, trained on a predefined number of word and character level n-grams, constructed using Scikit-learn's \emph{TfidfVectorizer} method. To find the best setting, we varied the SVM's C parameter in range $\{1,20,50,100,200\}$, the number of word features between $\{10000,50000, 100000,200000\}$ and character features between $\{0,30\}$\footnote{
In Figure~\ref{fig:appendix-ablation} (Appendix C), the reader can observe the results of the initial experiments on the number of word features that led to selection of this hyperparameter range.}.
Note that the word features were sorted by decreasing frequency. Here, we considered (word) n-grams of lengths between two and six. This SVM variation is from this point on referred to as ``SVM (generic)''. The main difference between ``SVM (generic)'' and ``SVM (Martinc et al.)'' is that the latter approach also considers punctuation-based and suffix-based features. Further, it is capable of constructing features that represent document sentiment, which was proven to work well for social media data sets (e.g., tweets). Finally, Martinc's approach also accounts for character repetitions and has a parameter for social-media text cleaning in preprocessing.
Note that for both SVM approaches we fine-tuned  the hyperparameter C, as is common when employing SVMs. The hyperparameter's values govern how penalized the learner is for a miss-classified instance, which is a property that was shown to vary across data sets (see for example \cite{MEYER2003169}).

\item[\textbf{Hierarchical attention networks (HILSTM).}]

The first neural network baseline is the recently introduced hierarchical attention network \cite{yang2016hierarchical}. Here, we performed a grid search over $\{64,128,256\}$ hidden layers sizes, embedding sizes of $\{128,256,512\}$, batch sizes of $\{8,24,52\}$ and number of epochs $\{5,15,20,30\}$. For detailed explanation of the architecture, please refer to the original contribution \cite{yang2016hierarchical}. We discuss the best-performing architecture in Section~\ref{sec:results} below.

\item[\textbf{Deep feedforward neural networks.}]
As tax2vec constructs feature vectors, we also attempted to use them as inputs for a standard feedforward neural network architecture \cite{lecun2015deep,schmidhuber2015deep}.
Here, we performed a grid search across hidden layer settings: $\{(128,64),(10,10,10)\}$ (where for example $(128,64)$ corresponds to a two hidden layer neural network, where in the first hidden layer there are 128 neurons and 64 in the second), batch sizes $\{8,24,52\}$  and the number of training epochs $\{5,15,20\}$.\footnote{The two deep architectures were implemented using TensorFlow \cite{45166}, and trained using a Nvidia Tesla K40 GPU. We report the best result for top 30 semantic features with the rarest terms heuristic.}

\end{description}
\subsection{Semantic features}
In addition to the semantic features constructed by tax2vec, doc2vec-based semantic features \cite{le2014distributed} were used as a baseline in order to allow for a simple comparison between two semantic feature construction approaches.  They were concatenated with the features constructed by Martinc et al.'s SVM approach described in Section \ref{subsec:classifiers}, in order to compare the benefits merging the BoW-based representations with a different type of semantic features (embedding-based ones). We set the embedding dimension to 256, as it was shown that lower dimensional embeddings do not perform well \cite{pennington2014glove}. 

\subsection{Description of the experiments}
The experiments were set up as follows. For the drug-related data sets, we used the splits given in the original paper \cite{GraBer}. For other data sets, we trained the classifiers using stratified $90\%:10\%$ splits. For each classifier, 10 such splits were obtained. The measure used in all cases is $F_1$, where for the multiclass problems (e.g., MBTI), we use the micro-averaged $F_1$. All experiments were repeated five times using different random seeds. The features obtained using tax2vec are used in combination with SVM classifiers, while the other classifiers are used as baselines.\footnote{ Note that simple feedforward neural networks could also be used in combination with hypernym features---we leave such computationally expensive experiments for further work.}

\section{Classification results}
\label{sec:results}
In this section we provide the results obtained by conducting the experiments outlined in the previous section. 
We begin by discussing the overall classification performance with respect to different heuristics used. Next, we discuss how tax2vec augments the learner's ability to classify when the number of text segments per user is reduced.

\subsection{Classification performance evaluation}

\begin{table}[b!]
    \centering
    \caption{Effect of the added semantic features to classification performance, where all text segments (tweets/comments per user or segments per news article) are used. 
The best performing feature selection heuristic for the majority of top performing classifiers was ``rarest terms'' or 
``Closeness centrality'', indicating that only a handful of hypernyms carry added value, relevant for classification. Note that the results 
in the table correspond to the best performing combination of a classifier and a given heuristic. }
        
    \resizebox{\textwidth}{!}{
\begin{tabular}{c|c|cccccc}
\# Semantic   & Learner & PAN (Age) & PAN (Gender) & MBTI & BBC News & Drugs (effect) & Drugs (side) \\ \hline \hline

  0   & HILSTM &                            0.422 &                               0.752 &              0.407 &                0.833 &                                  0.443 &                              0.514 \\
  0   & SVM (Martinc et al.) &                            0.417 &                               0.814 &             \fbox{ 0.682} &                0.983 &                                  0.468 &                                0.503 \\
    0   & SVM (generic) &                            0.424 &                               0.751 &              0.556 &                0.967 &                                  0.445 &                                0.462 \\
   256 (doc2vec)  & SVM (Martinc et al.) &                            0.422 &                               \fbox{0.817} &              0.675 &                0.979 &                                  0.416 &                                \fbox{0.523} \\
    30 (tax2vec) & DNN &                              0.400 &                               0.511 &              0.182 &                0.353 &                                    0.400 &                                0.321 \\
   \hline
10 (tax2vec)   & SVM (Martinc et al.) &                            0.445 &                               0.815 &              0.679 &              \fbox{  0.996} &                                  \fbox{ 0.47} &                                0.506 \\
     & SVM (generic) &                         \fbox{0.502} &                               0.781 &              0.556 &                0.972 &                                  0.445 &                                0.469 \\
25 (tax2vec)   & SVM (Martinc et al.) &                            0.454 &                               0.814 &              0.681 &                0.984 &                                  0.468 &                                  0.500 \\
     & SVM (generic) &                            0.484 &                               0.755 &              0.554 &                0.967 &                                  0.449 &                                0.466 \\
50 (tax2vec)   & SVM (Martinc et al.) &                            0.439 &                               0.814 &             0.681 &                0.983 &                                  0.462 &                                0.499 \\
     & SVM (generic) &                        0.444 &                               0.751 &              0.554 &                0.963 &                                  0.446 &                                0.463 \\
100 (tax2vec)  & SVM (Martinc et al.) &                            0.424 &                             0.816 &              0.678 &                0.984 &                                  0.466 &                                0.496 \\
     & SVM (generic) &                            0.422 &                               0.749 &              0.551 &                0.958 &                                  0.443 &                                 0.46 \\
500 (tax2vec)  & SVM (Martinc et al.) &                            0.383 &                               0.797 &              0.662 &                0.975 &                                   0.45 &                                0.477 \\
     & SVM (generic) &                              0.400 &                               0.724 &              0.532 &                0.909 &                                  0.424 &                                0.438 \\
1000 (tax2vec) & SVM (Martinc et al.) &                            0.368 &                               0.783 &              0.647 &                0.964 &                                  0.436 &                                0.466 \\
     & SVM (generic) &                            0.373 &                               0.701 &              0.512 &                0.851 &                                  0.407 &                                 0.420 \\
     \hline
\end{tabular}
}
    \label{tbl:f1}
\end{table}

The $F_1$ results are presented in Table~\ref{tbl:f1}. The first observation is that combining BoW-based representations with semantic features (tax2vec or doc2vec) leads to performance improvements in five out of six cases (MBTI being the only data set where no improvement is detected). Tax2vec outperforms doc2vec-based vectors in three out of five data sets (PAN 2016 (Age), BBC News and Drugs (effect)), while doc2vec-based features outperform tax2vec on two data sets (PAN 2017 (gender) and Drugs (Side)).


When it comes to tax2vec, up to 100 semantic features aid the SVM learners to achieve better accuracy. The most apparent improvement can be observed for the case of PAN 2016 (Age) data set, where the task was to predict age. Here, 10 semantic features notably improved the classifiers' performance (up to approximately $7\%$ for SVM (generic)). Further, a minor improvement over the state-of-the-art was also observed on the PAN 2017 (Gender) data set and the BBC news categorization (see results for SVM (Martinc et al.)). Hierarchical attention networks outperformed all other learners for the task of side effects prediction, yet semantics-augmented SVMs outperformed neural models when general drug effects were considered as target classes. Similarly, no performance improvements were offered by tax2vec on the MBTI data set.

\begin{figure}[b!]
    \centering
    \includegraphics[width=\linewidth]{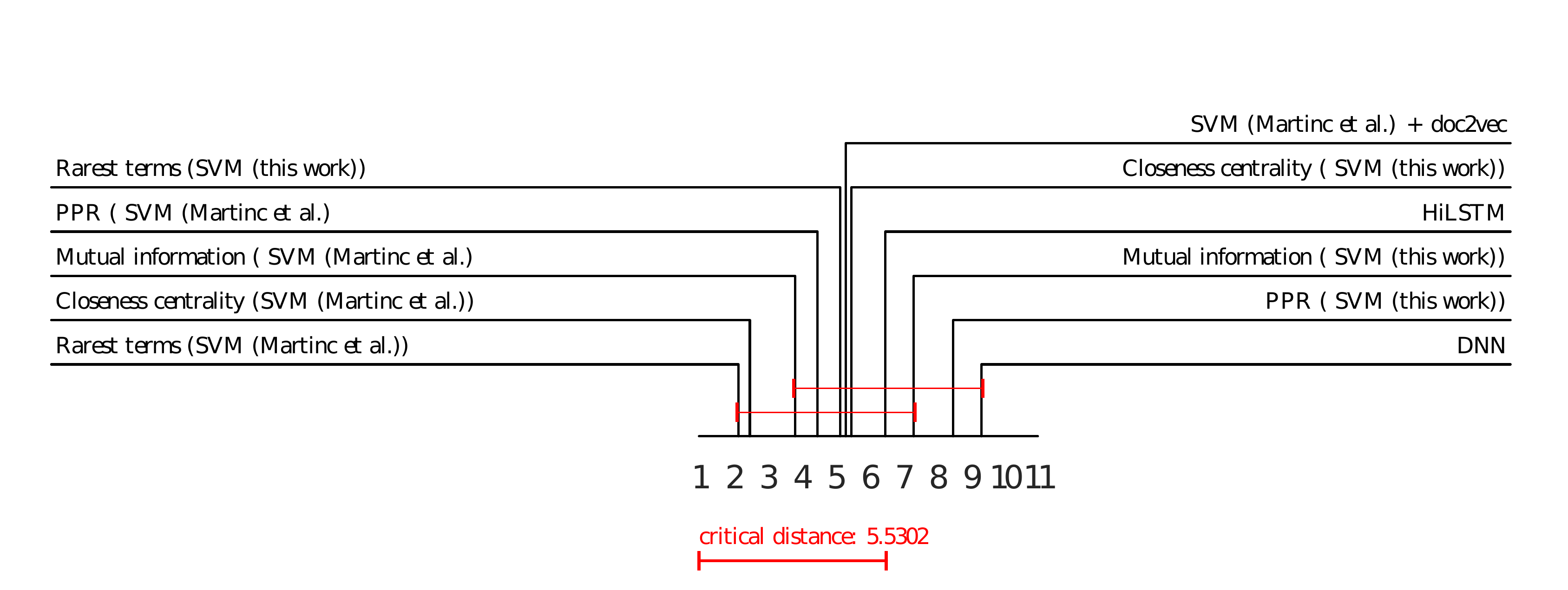}
    \caption{Average overall classifier ranks. The top (on average) performing classifier is an SVM (Martinc et al.) classifier augmented with semantic features, selected using  either simple frequency counts or closeness centrality.
    }
    \label{fig:all1}
\end{figure}

\begin{figure}[t!]
    \centering
    \includegraphics[width=.8\linewidth]{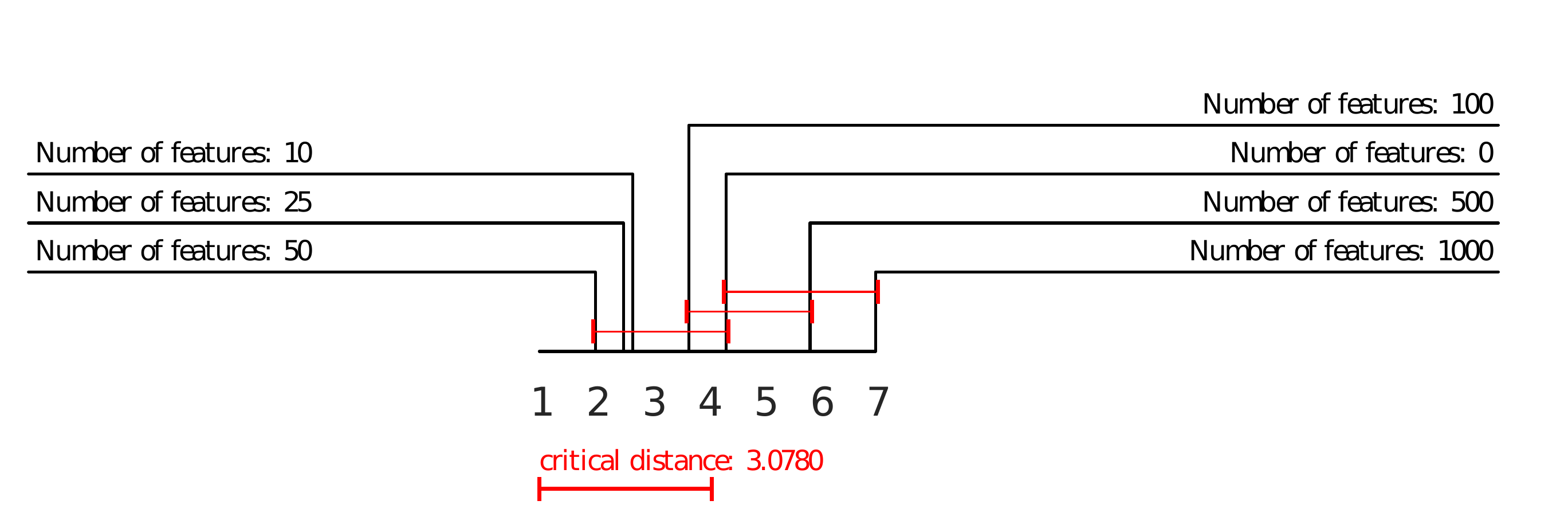}
    \caption{Effect of semantic features on average classifier rank. Up to 100 semantic features positively affects the classifiers' performance.}
    \label{fig:all2}
\end{figure}

\begin{figure}[t!]
\centering
        \includegraphics[width=.67\linewidth]{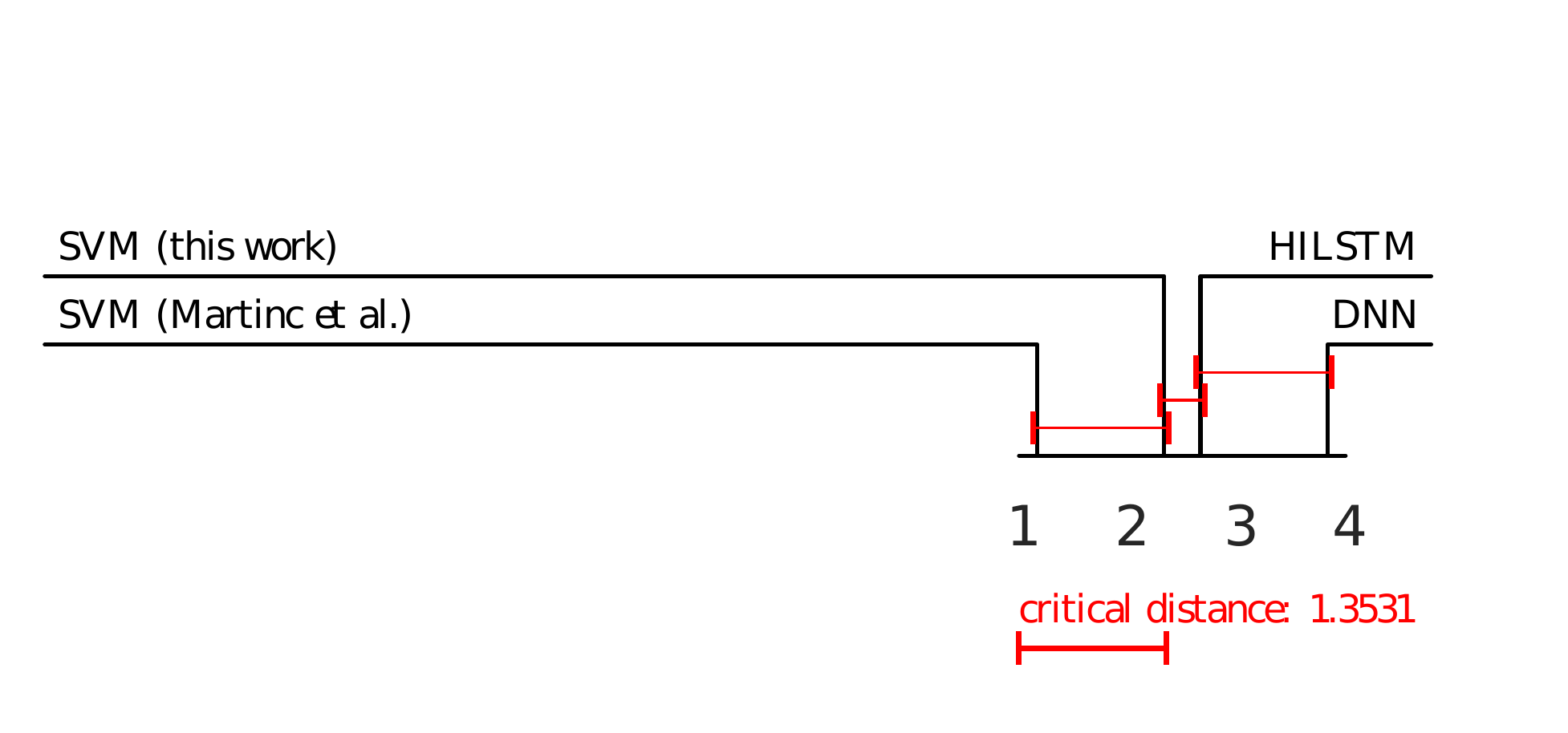}
    \caption{Overall model performance. SVMs dominate the short text classification. The diagram shows performance averaged over all data sets, where the best model parameterizations (see Table~\ref{tbl:f1}) were used for comparison.}
    \label{fig:all3}
\end{figure}

We now present the classification results in the form of critical distance diagrams, shown in Figures~\ref{fig:all1}, \ref{fig:all2} and \ref{fig:all3}.  The diagrams show average ranks of different algorithms according to the (micro) $F_1$ measure. A red line connects groups of classifiers that are not statistically significantly different from each other at a confidence level of $5\%$. The significance levels are computed using Friedman multiple test comparisons followed by Nemenyi post-hoc correction \cite{demvsar2006statistical}. For each data set, we selected the best performing parametrization (hyperparameter settings). The best (on average) performing C parameter for both SVM models was 50. The number of features that performed the best for all hyperparameter settings of the SVM (generic) considered in this study is 100{,}000. The HILSTM architecture's topology varied between data sets, yet we observed that the best results were obtained when more than 15 epochs of training were conducted, combined with the hidden layer size of 64 neurons, where the size of the attention layer was of the same dimension. 

In terms of feature selection, the following can be observed (Figure~\ref{fig:all1}). On average, the best performance was obtained when rarest terms heuristic was considered (first and fifth rank). Further, rarest terms, as well as the Personalized PageRank performed better (on average) than mutual information, which can be considered as a baseline in this comparison. The results indicate that myopic feature selection is not optimal when considering novel semantic features. We can also observe that on average the configuration with doc2vec semantic features (SVM (Martinc et al.) + doc2vec) performs worse (ranking as sixth) than all other configurations with SVM (Martinc et al.).

In Figure~\ref{fig:all2}, the reader can observe the performances \emph{of all learners}, averaged w.r.t. to the number of semantic features. The drawn diagram indicates that adding 10, 25 or 50 features to a classifier perform similarly well, however, as also discussed in the previous paragraph, the performance drops when larger semantic space is considered.

Finally, in Figure~\ref{fig:all3} it can be observed that the overall performance of Martinc et al.'s SVMs is the best, followed by generic SVMs, as well as HILSTMs. We believe such performance drop with deep neural networks in general is due to concatenation of documents prior to learning, and as only a fixed sequence length can be considered, potentially large parts of the token space were neglected during learning. A similar result was, for example observed in the most recent PAN competition \cite{martinc2019fake}. 

\subsection{Few-shot (per instance) learning}
As discussed in the introductory sections, one of the goals of this paper was also to explore the setting, where only a handful of text segments per user are considered. Even though such setting is not strictly a few-shot learning \cite{snell2017prototypical}, reducing the number of text segments per instance (e.g., user) aims to simulate a setting where there is limited information available. In Table~\ref{tab:oneshot}, we present the results for the setting, where only (up to) 10 text segments (e.g., tweets or paragraphs in a given news article) were used for training.

The segments were sampled randomly. Only a single text segment per user was considered for the medical texts, as they consist of at max of three commentaries. Similarly, as the BBC news data set consists of news article-genre pairs, we split the news article to paragraphs, which we randomly sampled. The rationale for such sampling is to be able to evaluate tax2vec's performance when, for example, only a handful of paragraphs are available (e.g., only the lead). 

\begin{table}[t]
    \centering
    \caption{Effect of added semantic features to classification performance---few shot learning.}
    \resizebox{\textwidth}{!}{
\begin{tabular}{c|c|cccccc}

 Semantic  (tax2vec)   & Learner & PAN (Age) & PAN (Gender) & MBTI & BBC News & Drugs (effect) & Drugs (side) \\ \hline \hline
0    & SVM (Martinc et al.) &                            0.378 &                               0.617 &              0.288 &                0.977 &                                  0.468 &                                0.503 \\
     & SVM (generic) &                            0.429 &                               0.554 &              0.225 &                0.936 &                                  0.445 &                                0.462 \\ \hline
10   & SVM (Martinc et al.) &                             0.39 &                               0.616 &             \fbox{ 0.292 } &              \fbox{  0.981 } &                                   0.47 &                                0.503 \\
     & SVM (generic) &                            0.429 &                               0.557 &              0.225 &                0.948 &                                  0.444 &                                0.464 \\
25   & SVM (Martinc et al.) &                            0.429 &                             \fbox{  0.618 }&              0.288 &                0.979 &                                  0.465 &                                  0.5 \\
     & SVM (generic) &                         \fbox{  0.439 }&                               0.562 &              0.226 &                0.933 &                                  0.445 &                                0.458 \\
50   & SVM (Martinc et al.) &                            0.402 &                               0.617 &              0.288 &                0.974 &                                  0.474 &                               \fbox{ 0.504 } \\
     & SVM (generic) &                             0.42 &                               0.557 &              0.225 &                0.919 &                                  0.442 &                                 0.46 \\
100  & SVM (Martinc et al.) &                            0.382 &                               0.614 &              0.286 &                0.974 &                                \fbox{ 0.476 } &                                0.493 \\
     & SVM (generic) &                            0.411 &                               0.552 &              0.223 &                0.906 &                                  0.437 &                                0.457 \\
500  & SVM (Martinc et al.) &                            0.359 &                               0.604 &              0.276 &                0.959 &                                  0.465 &                                0.471 \\
     & SVM (generic) &                            0.365 &                               0.548 &               0.22 &                  0.8 &                                  0.419 &                                0.435 \\
1000 & SVM (Martinc et al.) &                             0.34 &                                0.59 &              0.266 &                0.925 &                                  0.442 &                                 0.46 \\
     & SVM (generic) &                            0.359 &                               0.535 &              0.213 &                0.704 &                                  0.412 &                                0.417 \\
     \hline
\end{tabular}
}
    \label{tab:oneshot}
\end{table}

We observe that tax2vec based features improve the learners' performance on all of the data sets, albeit by a small margin. The results indicate that adding semantic information improves the performance as only a handful of text segments does not necessarily contain the relevant information.

\subsection{Few-shot learning results}
\label{sec:fsl}
We next discuss the results of few-shot learning, as to our knowledge this type of experiments were not conducted before in combination with semantic feature construction methods. The first observation is, semantic features indeed offer more consistent performance improvements than those observed in Table~\ref{tab:hyp2}, where incremental improvements were not observed on all data sets. In a few-shot learning scenario, however, on all data sets, the inclusion of semantic space either resulted in similar or better performance, indicating a consistent positive effect on the learning in a limited setting. The differences in learner's performance vary around 1\% improvement. For example, a ~1\% improvement was observed for PAN 2016 (Age), BBC News and MBTI data sets. 

We finally comment on the classification performance when considering the BBC data set when comparing to reported state-of-the-art. The observed results ($\geq$98\%) are competitive to neural approaches, such as for example as reported in \cite{asim2019robust}, where similar span of accuracy was observed. Furthermore, doc2vec-based models have been observed to perform similarly \cite{trieu2017news}. The results of this work indicate that by considering smaller number of paragraphs (instead of whole documents), competitive performance can be observed on the BBC data set.


\subsection{Interpretation of results}
\label{sec:inter}
In this section we explain the intuition behind the effect of semantic features on the classifier's performance.
Note that the best performing SVM models consisted of thousands of tf-idf word and character level features, yet only up to 100 semantic features, when added, notably improved the performance. This effect can be understood via the way SVMs learn from high-dimensional data. With each new feature, we increase the dimensionality of the feature space. Even a single feature, when added, potentially impacts the hyperplane construction. Thus, otherwise problem-irrelevant features can become relevant when novel features are added. We believe that adding semantic features to 
(raw) word tf-idf vector space introduces new information, crucial for successful learning, and potentially aligns the remainder of features so that the classifier can better separate the points of interest. 

The other explanation for the notable differences in predictive performance is possibly related to small data set sizes, where only a handful of features can be of relevance and thus notably impact a given classifier's performance. We next discuss the impact of the number of selected semantic features on performance.

\subsection{How large semantic space should be considered?}
\label{sec:sspace}
Tables~\ref{tab:oneshot} and \ref{tab:hyp2} show that a relatively small number of semantic features are needed for potential performance gains.
Note that the number of semantic features that need to be considered is around $\leq 100$ in most of the cases. The results indicate that a relatively small proportion of the semantic space carries relevant (additional) information, whereas the remainder potentially introduces noise that degrades the performance. Note that in the limit every term from the taxonomy derived from a given corpus could be considered. In such a scenario, many terms would be irrelevant and would only introduce noise. The experiments conducted in this paper indicate that the threshold for the number of features is in the order of hundreds, yet not more features. 

\section{Qualitative assessment and explainability of tax2vec}
\label{sec:qual+inter}

This section discusses the properties of the resulting semantic space in Section~\ref{sec:qual}, which is followed by a discussion on the explainability of the proposed tax2vec algorithm in Section~\ref{sec:explain}.

\subsection{Analysis of the resulting semantic space}
\label{sec:qual}
In this section we discuss the qualitative properties of the obtained corpus-based taxonomies. We present the results concerning hypernym frequency distributions, as well as the overall structure of an example corpus-based taxonomy.

As the proposed approach is entirely symbolic---each feature can be traced back to a unique hypernym---we explored the feature space qualitatively by exploring the statistical properties of the induced taxonomy using graph-statistical approaches. Here, we modeled hypernym frequency distributions to investigate possible similarity with the Zipf's law \cite{piantadosi2014zipf}. The analysis was performed using the Py3plex library \cite{10.1007/978-3-030-05411-3_60}. We also  visualized the document-based taxonomy of the PAN 2016 (Age) data set using Cytoscape \cite{shannon2003cytoscape}. 

\begin{figure}[ht]
\centering
\includegraphics[width=0.8\linewidth]{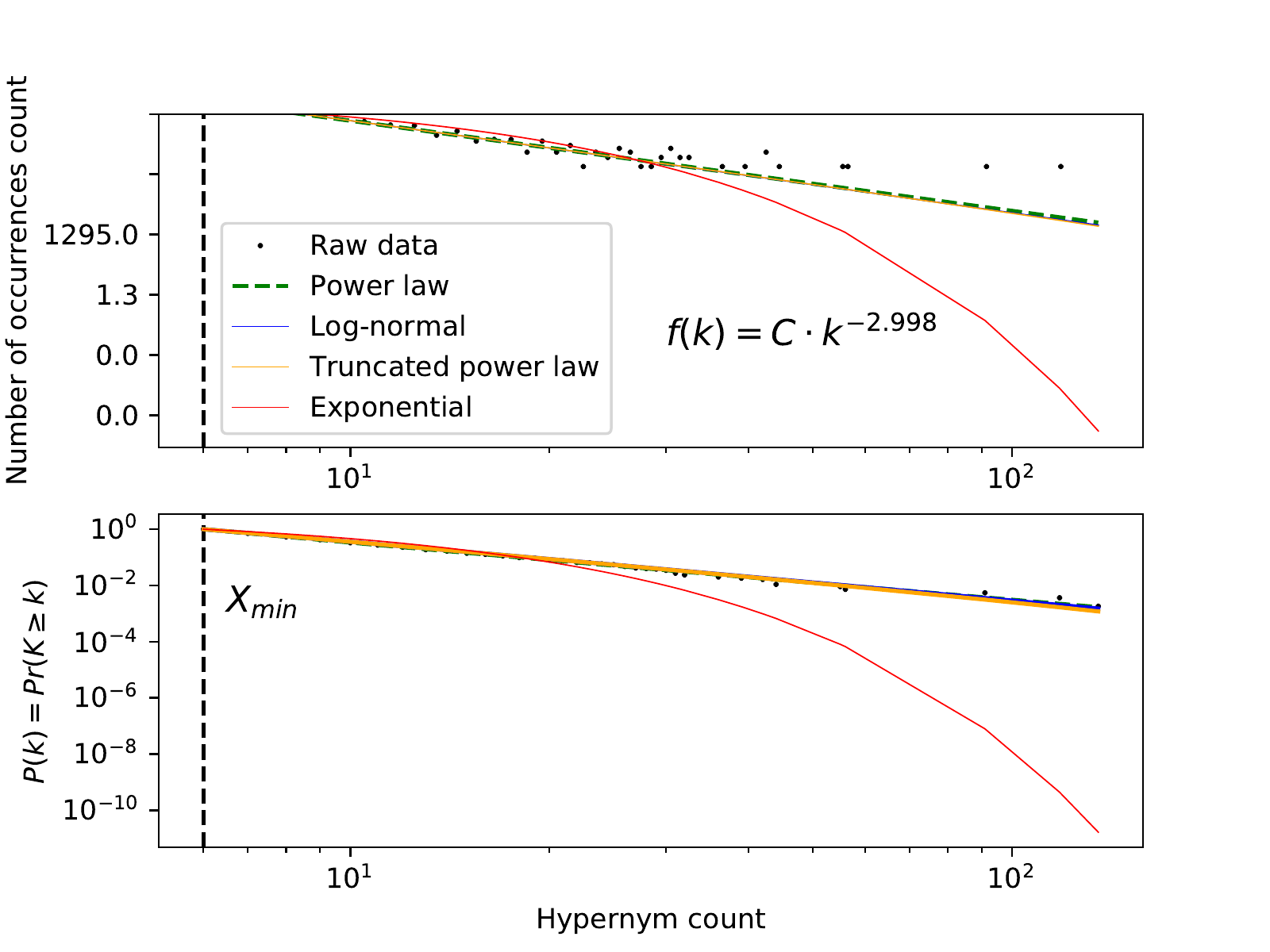}
\caption{Hypernym frequency distribution for the PAN 2016 (Age) data set. The equation above the upper plot denotes the coefficients of a power law distribution ($C$ is a constant). In real world phenomena, the exponent of the rightmost expression was observed to range between $\approx 2 $ and $ \approx 3$, indicating the hypernym structure of the feature space is subject to a heavy-tailed (possibly best fit---power law) distribution. The $X_{min}$ denotes the hypernym count, after which notable differences in hypernym counts---scale free behavior is observed. Such distribution is to some extent expected, as some hypernyms are more general than others, and thus present in more document-hypernym mappings.}
\label{fig:freq}
\end{figure}

The examples in this section are all based on the corpus-based taxonomy, constructed from the PAN 2016 (Age) data set. The results of fitting various heavy-tailed distributions to the hypernym frequencies are given in Figure~\ref{fig:freq}.

\begin{figure}[t!]
\centering
\includegraphics[width=1\linewidth]{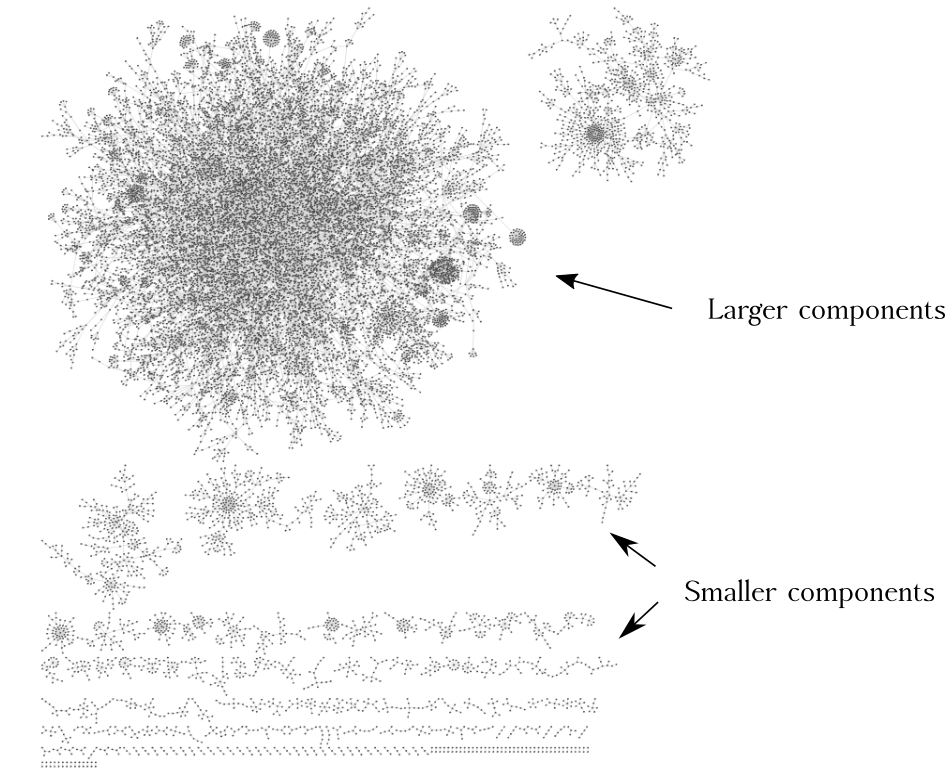}
\caption{Topological structure of the hypernym space, induced from the PAN 2016 (Age) data set. Multiple connected components emerged, indicating not all hypernyms map to the same high-level concepts. Such segmentation is data set-specific, and can also potentially provide the means to compare semantic spaces of different data sets. It can be observed that the obtained space is organized in multiple separate components. The largest are drawn at the topmost part of the figure, whereas the smaller ones at the bottom. Such segmentation corresponds to generalizations based on different parts of speech, e.g., nouns and verbs.}
\label{fig:hyp}
\end{figure}

We fitted power law, truncated normal, log-normal and exponential distributions to the hypernym frequency data. For detailed overview of the distributions we refer the reader to \cite{foss2011introduction}. One of the key properties we researched was whether the underlying hypernym distribution is exponential or not, as non-exponential distributions indicate similarity with the well known Zipf's law \cite{piantadosi2014zipf}. 
The hypernym corpus-based taxonomy is visualized in Figure~\ref{fig:hyp}.

Here, each node represents a hypernym obtained in word-to-hypernym mapping phase of tax2vec. The edges represent the hypernymy relation between a given pair of hypernyms.

We next present the results of modeling the corpus-based hypernym frequency distributions.
The two functions representing the best fit to hypernym frequency distributions are indeed the power law and the truncated power law. As similar behavior is observed for word frequency in documents \cite{piantadosi2014zipf}, we believe hypernym distributions are a natural extension, as naturally, if a high-frequency word maps to a given hypernym, the hypernym will be relatively more common with respect to the  occurrence of other hypernyms.

We observe that multiple connected components of varying sizes emerge. There exists only a single largest connected component, which consists of more general noun hypernyms, such as \emph{entity} and similar. Interestingly, many smaller components also emerged, indicating parts of the word vector space could be mapped to very specific, disconnected parts of the WordNet taxonomy. Some examples of small disconnected components include (one component per line), indicating also verb-level semantics can be captured and taken into account:

\begin{center}
\fbox{
\begin{minipage}{29em}
${'spot.v.02', 'discriminate.v.03'}
{'homestead.v.01', 'settle.v.21'}\\
{'smell.v.05', 'perceive.v.02', 'understand.v.02'}\\
{'dazzle.v.01', 'blind.v.01'}\\
{'romance.v.02', 'adore.v.01', 'care\_for.v.02', 'love.v.03', 'love.v.01'}\\
{'surrender.v.01', 'yield.v.12', 'capitulate.v.01'}$
\end{minipage}}
\end{center}

\subsection{Explainability of tex2vec}
\label{sec:explain}
As discussed in the previous sections, tax2vec selects a set of hypernyms according to a given heuristic and uses them for learning. One of the key benefits of such approach is that the selected semantic features can easily be inspected, hence potentially offering interesting insights into the semantics, underlying the problem at hand.

We discuss here a set of 30 features which emerged as relevant according to the ``mutual information'' heuristic when the BBC News and PAN 2016 (Age) data sets were considered. Here, tax2vec was trained on 90\% of the data, the rest was removed (test set). The features and their corresponding mutual information scores are shown in Table~\ref{tab:hyp2}.
 
\begin{table}[h]
\caption{Most informative features with respect to the target class (ranked by MI)---Classes represent news topics (BBC) and different age intervals (PAN 2016 (Age)). Individual target classes are sorted according to a descending mutual information with respect to a given feature.}
  \resizebox{\textwidth}{!}{
\begin{tabular}{c||cccccc}
& & \multicolumn{5}{c} {Sorted target class-mutual information pairs}\\
Semantic feature & Average MI & Class 1 & Class 2 & Class 3 & Class 4 & Class 5 \\ \hline \hline
BBC News data set \\
\hline
tory.n.03                      & 0.057 & politics:0.14      & entertainment:0.05 &  business:0.03      & sport:0.01  & x\\
movie.n.01                     & 0.059 & business:0.14      & politics:0.04      & entertainment:0.04 & sport:0.02 & x \\
conservative.n.01              & 0.061 & politics:0.15      & entertainment:0.05 & business:0.03      & sport:0.01 & x \\
vote.n.02                      & 0.061 & business:0.15      & entertainment:0.04 & politics:0.04      & sport:0.02 & x \\
election.n.01                  & 0.063 & entertainment:0.16 & business:0.05      & politics:0.04      & sport:0.0 & x \\
topology.n.04                  & 0.063 & entertainment:0.16 & business:0.05      & politics:0.04      & sport:0.0 & x \\
mercantile\_establishment.n.01 & 0.068 & politics:0.17      & business:0.07      & entertainment:0.03 & sport:0.01 & x \\
star\_topology.n.01            & 0.069 & politics:0.17      & business:0.07      & entertainment:0.03 & sport:0.01 & x \\
rightist.n.01                  & 0.074 & politics:0.18      & business:0.06      & entertainment:0.04 & sport:0.01 & x \\
marketplace.n.02               & 0.087 & entertainment:0.22 & business:0.06      & politics:0.05      & sport:0.01 & x \\ \hline
PAN (Age) data set \\ \hline
hippie.n.01              & 0.007 & 25-34:0.01 & 35-49:0.01 & 18-24:0.0  & 65-xx:0.0  & 50-64:0.0 \\
ceremony.n.03            & 0.007 & 25-34:0.01 & 35-49:0.01 & 18-24:0.01 & 65-xx:0.0  & 50-64:0.0 \\
resource.n.02            & 0.008 & 50-64:0.02 & 18-24:0.01 & 25-34:0.0  & 65-xx:0.0  & 35-49:0.0 \\
draw.v.07                & 0.008 & 25-34:0.02 & 35-49:0.01 & 50-64:0.01 & 65-xx:0.0  & 18-24:0.0 \\
observation.n.02         & 0.008 & 25-34:0.02 & 35-49:0.01 & 50-64:0.01 & 65-xx:0.0  & 18-24:0.0 \\
wine.n.01                & 0.008 & 35-49:0.02 & 25-34:0.01 & 18-24:0.01 & 50-64:0.01 & 65-xx:0.0 \\
suck.v.02                & 0.008 & 25-34:0.02 & 50-64:0.02 & 35-49:0.0  & 65-xx:0.0  & 18-24:0.0 \\
sleep.n.03               & 0.008 & 25-34:0.02 & 50-64:0.02 & 35-49:0.0  & 65-xx:0.0  & 18-24:0.0 \\
recognize.v.09           & 0.009 & 25-34:0.02 & 35-49:0.02 & 18-24:0.0  & 50-64:0.0  & 65-xx:0.0 \\
weather.v.04             & 0.009 & 25-34:0.02 & 50-64:0.02 & 35-49:0.0  & 18-24:0.0  & 65-xx:0.0 \\
invention.n.02           & 0.009 & 25-34:0.02 & 35-49:0.01 & 18-24:0.01 & 50-64:0.0  & 65-xx:0.0 \\
yankee.n.03              & 0.01  & 50-64:0.02 & 18-24:0.01 & 25-34:0.01 & 35-49:0.0  & 65-xx:0.0 \\
\hline
\end{tabular}
}
\label{tab:hyp2}
\end{table}

We can observe that the ``sport'' topic (BBC data set) is not well associated with the prioritized features. On the contrary, terms such as ``rightist'' and  ``conservative'' emerged as relevant for classifying into the ``politics'' class. Similarly, ``marketplace'' for example, appeared relevant for classifying into the ``entertainment'' class. Even more interesting associations emerged when the same feature ranking was conducted on the PAN 2016 (Age) data set. Here, terms such as  ``resource'' and ``wine'' were relevant for classifying middle-aged (``wine'') and older adult (``resource'') populations. Note that the older population (65-xx class) was not associated with any of the hypernyms. We believe the reason for this is that the number of available tweets decreases with age.
 
We repeated a similar experiment (BBC data set) using the ``rarest terms'' heuristic. The terms which emerged are:

\begin{center}
\fbox{
\begin{minipage}{32em}
'problem.n.02', 'question.n.02', 'riddle.n.01', 'salmon.n.04', 'militia.n.02', 'orphan.n.04', 'taboo.n.01', 'desertion.n.01', 'dearth.n.02', 'outfitter.n.02', 'scarcity.n.01', 'vasodilator.n.01', 'dilator.n.02', 'fluoxetine.n.01', 'high blood pressure.n.01', 'amlodipine besylate.n.01', 'drain.n.01', 'imperative mood.n.01', 'fluorescent.n.01', 'veneer.n.01', 'autograph.n.01', 'oak.n.02', 'layout.n.01', 'wall.n.01', 'firewall.n.03', 'workload.n.01', 'manuscript.n.02', 'cake.n.01', 'partition.n.01', 'plasterboard.n.01'
\end{minipage}}
\end{center}

Even if the feature selection method is unsupervised (not directly associated to classes), we can immediately observe that the features correspond to different topics, raging from medicine (e.g., “high blood presure”), politics (e.g., “militia”), food(e.g., “cake”) and more, indicating that the rarest hypernyms are indeed diverse and as such potentially useful for the learner.

The results suggest that tax2vec could potentially also be used to inspect the semantic background of a given data set directly, regardless of the learning task. We believe there are many potential uses for the obtained features,  including the following, to be addressed in further work.
\begin{itemize}
\item Concept drift detection, i.e. topics change over time; could it be qualitatively detected?
\item Topic domination, i.e. what type of topic is dominant with respect to e.g., a geographical region inspected?
\item What other learning tasks can benefit by using second level semantics? Can the obtained features be used, for example, for fast keyword search?
\end{itemize}

\section{Implementation and availability}
\label{sec:availability}
The tax2vec algorithm is implemented in Python 3, where Multiprocessing\footnote{\url{https://docs.python.org/2/library/multiprocessing.html}}, SciPy \cite{scipy} and Numpy \cite{walt2011numpy} libraries are used for fast (sparse), vectorized operations and parallelism. 

As performing a grid search over several parameters is computationally expensive, the majority of the experiments were conducted using the SLING supercomputing architecture.\footnote{\url{http://www.sling.si/sling/}}

We developed a stand-alone library that relatively seamlessly  fits into existing text mining workflows, hence the Scikit-learn's model syntax was adopted \cite{pedregosa2011scikit}. The algorithm is first initiated as an object:
\begin{equation*}
\boxed{
\textrm{vectorizer} = \textrm{tax2vec(heuristic,number of features)}
}
\end{equation*}
\noindent followed by standard \emph{fit} and \emph{transform} calls:
\begin{equation*}
\boxed{
\textrm{new\_features} = \textrm{vectorizer}.\textrm{fit\_transform(corpus, optional labels)}
}
\end{equation*}

Such implementation offers fast prototyping capabilities, needed ubiquitously in the development of learning algorithms and executable NLP and text mining workflows. 

The proposed tax2vec approach is freely available as a Python 3 library at \url{https://github.com/SkBlaz/tax2vec}, which includes also the installation instructions.

\section{Conclusions and future work}
\label{sec:conclusions}
In this work we propose tax2vec, a parallel algorithm for taxonomy-based enrichment of text documents. Tax2vec first maps the words from individual documents to their hypernym counterparts, which are considered as candidate features and weighted according to a normalized tf-idf metric. To select only a user-specified number of relevant features, tax2vec implements multiple feature selection heuristics, which select only the potentially relevant features. The sparse matrix of constructed features is finally used alongside the bag-of-words document representations for the task of text classification, where we study its performance on small data sets, where both the number of text segments per user, as well as the number of overall users considered are small.

The tax2vec approach considerably improves the classification performance especially on data sets consisting of tweets, but also on the news. The proposed implementation offers a simple-to-use API, which facilitates inclusion into existing text preprocessing workflows.

As the next step, the tax2vec will be tested on SMS spam data \cite{delany2012sms}, which is another potentially interesting short text data set where taxonomy-based features could improve performance and help the user better understand what classifies as spam (and what not).

One of the drawbacks we plan to address is the support for arbitrary directed acyclic multigraphs---structures commonly used to represent background knowledge. Support for such knowledge would offer a multitude of applications in e.g., biology, where gene ontology and other resources which annotate entities of interest are freely available.

In this work we focus on BoW representation of documents, yet we believe tax2vec could also be used along Continuous Bag-of-Words (CBoW) models. We leave such experimentation for further work.

Even though we use Lesk for the disambiguation task, we believe recent advancements in neural disambiguation \cite{iacobacci2016embeddings} could also be a ``drop-in'' replacement for this part of tax2vec. We leave the exploration of such options for further work.

In this work we explored how WordNet could be adapted for scalable feature construction, however tax2vec is by no means limited to manually curated relational (hierarchical) structures. As part of the further work, we believe feature construction based on \emph{knowledge graphs} could also be an option.

The abundance of neural embedding methods introduced in the recent years can be complementary to tax2vec. Understanding how the performance can be improved by jointly using both tax2vec's features and neural network-based ones is a potential interesting research opportunity.  Further, in NLP setting, not much attention is devoted to this topic, thus we believe these results offer new trajectories for few-shot learning research.

Other further work considers joining the tax2vec features with existing state-of-the-art deep learning approaches, such as the hierarchical attention networks, which are---according to this study---not very suitable for learning on scarce data sets. We believe that the introduction of semantics into deep learning could be beneficial for both performance, as well as the  interpretability of currently poorly understood black-box models.

Finally, as the main benefit of tax2vec is its explanatory power, we believe it could be used for fast keyword 
search; here, for example, new news or articles could be used as inputs, where the ranked list of semantic features could be directly used as candidate keywords.

\section*{Acknowledgements}
We would first like to thank the reviewers for insightful comments that improved this paper.
The work of the first author was funded by the Slovenian Research Agency through a young researcher grant.
The work of other authors was supported by the Slovenian Research Agency (ARRS) core research programme \emph{Knowledge Technologies} (P2-0103), an ARRS funded research project \emph{Semantic Data Mining for Linked Open Data} (financed under the ERC Complementary Scheme, N2-0078) and European Union\'s Horizon 2020 research and  innovation programme under grant agreement No 825153, project EMBEDDIA (Cross-Lingual Embeddings for
Less-Represented Languages in European News Media). This research has received funding from the European Union’s Horizon 2020 research and innovation programme under grant agreement No 769661, SAAM project. We also gratefully acknowledge the support of NVIDIA Corporation for the donation of  Titan-XP GPU.

\bibliography{bibliography}
\appendix

\section{Personalized PageRank algorithm}
\label{sec:appendix-ppr}
The Personalized PageRank (PPR) algorithm is described below.
Let $V$ represent the nodes of the corpus-based taxonomy.
For each node $u \in V$, a feature vector is computed by calculating the stationary distribution of a random walk, starting at node $u$. The stationary distribution is approximated by using power iteration, where the $i$-th component of the approximation in the $k$-th iteration is computed as
\begin{equation}
\label{eqPR}
  \gamma_{u}(i)^{(k+1)} = \alpha \cdot \sum_{j \rightarrow i}\frac{\gamma_{u}(j)^{(k)}}{d_{j}^{out}}+(1-\alpha) \cdot v_{u}(i);k = 1, 2,\dots
\end{equation}
\noindent The number of iterations $k$ is increased until the stationary distribution converges to the stationary distribution vector (PPR value for node $i$). 
In the above equation, $\alpha$ is the damping factor that corresponds to the probability that a random walk follows a randomly chosen outgoing edge from the current node rather than restarting its walk. The summation index $j$ runs over all nodes of the network that have an outgoing connection toward $j$, (denoted as $j \rightarrow i$ in the sum), and $d_{j}^{out}$ is the out degree of node $d_{j}$. The term $v_{u}(i)$ is the restart distribution that corresponds to a vector of probabilities for a walker's return to the starting node $u$, i.e. $v_{u}(u) = 1$ and $v_u(i)=0$ for $i\neq u$. This vector guarantees that the walker will jump back to the starting node $u$ in case of a restart.\footnote{Note that if the binary vector were instead composed exclusively of ones, the iteration would compute the  global PageRank vector, and Equation~\ref{eqPR} would correspond to the standard PageRank iteration.}

\section{Example document split}
\label{sec:appendix-splits}
While for the data sets consisting of tweets and short comments, the number of segments in a document corresponds to the number of tweets or comments by a user, in the news data set, we varied the size of the news (to create short documents) by splitting the news into paragraphs (we denote paragraph splits with |||). See example segmentation of news from BBC data set\footnote{\url{https://github.com/suraj-deshmukh/BBC-Dataset-News-Classification/blob/master/dataset/dataset.csv}} below.

||| The decision to keep interest rates on hold at 4.75\% earlier this month was passed 8-1 by the Bank of England's rate-setting body, minutes have shown.||| One member of the Bank's Monetary Policy Committee (MPC) - Paul Tucker - voted to raise rates to 5\%. The news surprised some analysts who had expected the latest minutes to show another unanimous decision. Worries over growth rates and consumer spending were behind the decision to freeze rates, the minutes showed. The Bank's latest inflation report, released last week, had noted that the main reason inflation might fall was weaker consumer spending.||| However, MPC member Paul Tucker voted for a quarter point rise in interest rates to 5\%. He argued that economic growth was picking up, and that the equity, credit and housing markets had been stronger than expected.||| The Bank's minutes said that risks to the inflation forecast were ``sufficiently to the downside'' to keep rates on hold at its latest meeting. However, the minutes added: ``Some members noted that an increase might be warranted in due course if the economy evolved in line with the central projection''. Ross Walker, UK economist at Royal Bank of Scotland, said he was surprised that a dissenting vote had been made so soon. He said the minutes appeared to be ``trying to get the market to focus on the possibility of a rise in rates''. ``If the economy pans out as they expect then they are probably going to have to hike rates.'' However, he added, any rate increase is not likely to happen until later this year, with MPC members likely to look for a more sustainable pick up in consumer spending before acting.

This news article is split by a parser into the following four segments (and in short document setting only one paragraph is used to represent the document).

\begin{itemize}
\item The decision to keep interest rates on hold at 4.75\% earlier this month was passed 8-1 by the Bank of England's rate-setting body, minutes have shown.
\item One member of the Bank's Monetary Policy Committee (MPC) - Paul Tucker - voted to raise rates to 5\%. The news surprised some analysts who had expected the latest minutes to show another unanimous decision. Worries over growth rates and consumer spending were behind the decision to freeze rates, the minutes showed. The Bank's latest inflation report, released last week, had noted that the main reason inflation might fall was weaker consumer spending.
\item However, MPC member Paul Tucker voted for a quarter point rise in interest rates to 5\%. He argued that economic growth was picking up, and that the equity, credit and housing markets had been stronger than expected.
\item The Bank's minutes said that risks to the inflation forecast were``sufficiently to the downside'' to keep rates on hold at its latest meeting. However, the minutes added: ``Some members noted that an increase might be warranted in due course if the economy evolved in line with the central projection.'' Ross Walker, UK economist at Royal Bank of Scotland, said he was surprised that a dissenting vote had been made so soon. He said the minutes appeared to be ``trying to get the market to focus on the possibility of a rise in rates.'' ``If the economy pans out as they expect then they are probably going to have to hike rates.'' However, he added, ``any rate increase is not likely to happen until later this year, with MPC members likely to look for a more sustainable pick up in consumer spending before acting.''
\end{itemize}

\section{Impact of different number of features across data sets}
\label{sec:appendix-ablation}
\begin{figure}[h!]
    \centering
    \includegraphics[width = \linewidth]{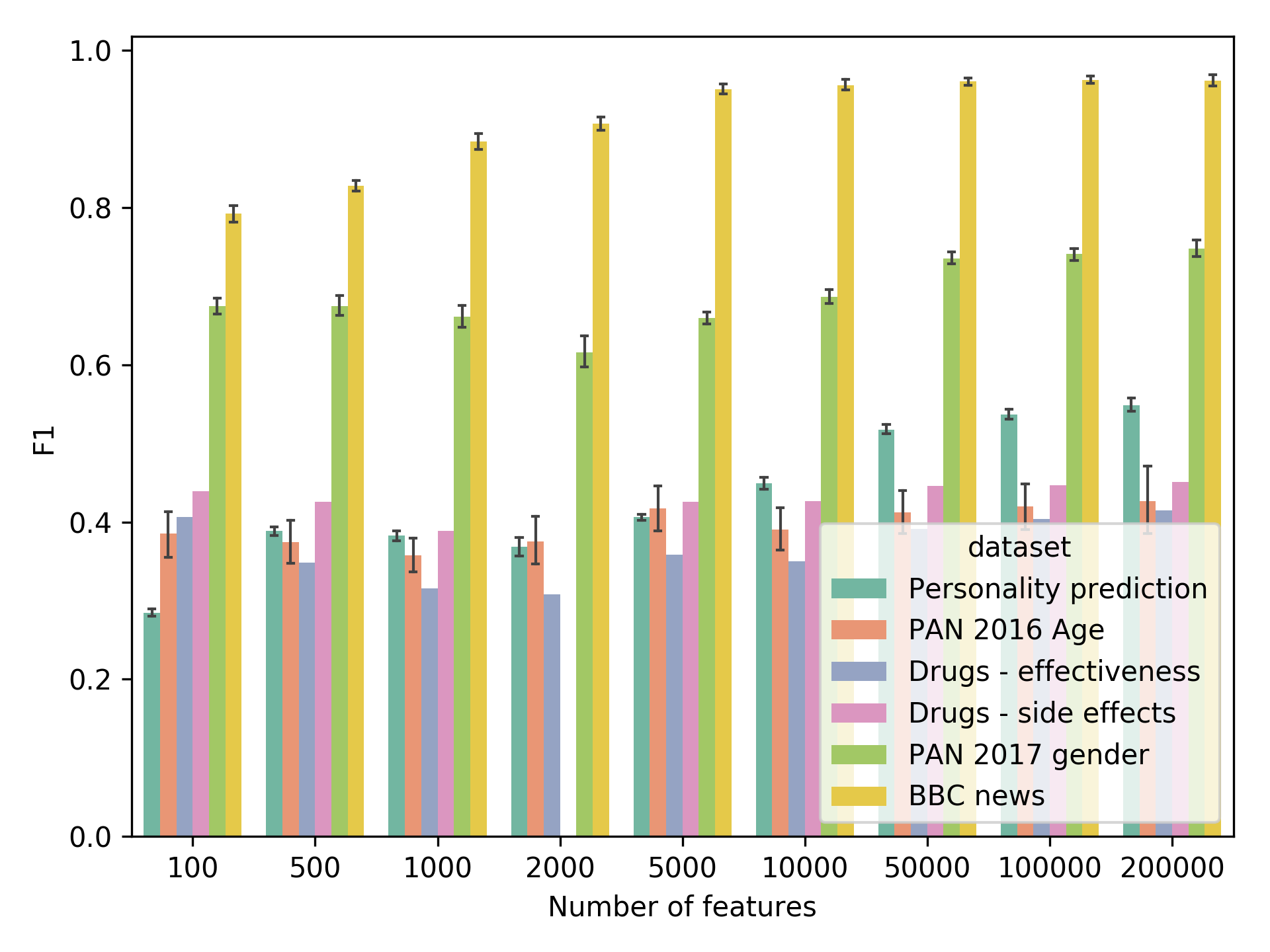}
    \caption{Impact of the number of features used by the SVM (generic) 
    on the F1 performance. The best performances were observed for feature numbers (word tokens) $\geq$ 10{,}000, hence these feature numbers were considered in the more expensive experiment stage with semantic vectors.}
    \label{fig:appendix-ablation}
\end{figure}

\end{document}